# A Lightweight Force-Controllable Wearable Arm Based on Magnetorheological-Hydrostatic Actuators

Catherine Véronneau[1], Jeff Denis[1], Louis-Philippe Lebel[1], Marc Denninger[2], Jean-Sébastien Plante[2], Alexandre Girard[1]
*Abstract*— Supernumerary Robotic Limbs (SRLs) are wearable robots augmenting human capabilities by acting as a co-worker, reaching objects, support human arms, etc. However, existing SRLs lack the mechanical backdrivability and bandwidth required for tasks where the interaction forces must be controllable such as painting, manipulating fragile objects, etc. Being highly backdrivable with a high bandwidth while minimizing weight presents a major technological challenge imposed by the limited performances of conventional electromagnetic actuators. This paper studies the feasibility of using magnetorheological (MR) clutches coupled to a low-friction hydrostatic transmission to provide a highly capable, but yet lightweight, force-controllable SRL. A 2.7 kg 2-DOFs wearable robotic arm is designed and built. Shoulder and elbow joints are designed to deliver 39 and 25 Nm, with 115 and 180° of range of motion. Experimental studies conducted on a one-DOF test bench and validated analytically demonstrate a high force bandwidth (>25 Hz) and a good ability to control interaction forces even when interacting with an external impedance. Furthermore, three force-control approaches are studied and demonstrated experimentally: open-loop, closed-loop on force, and closed-loop on pressure. All three methods are shown to be effective. Overall, the proposed MR-Hydrostatic actuation system is well-suited for a lightweight SRL interacting with both human and environment that add unpredictable disturbances.

## I. INTRODUCTION

In a context of aging workforce and urbanization, a lot of economic sectors such as agriculture, construction, and manufacturing suffer from a shortage of workers [1]. Moreover, the complexity and fatiguing nature of industrial assembly tasks result in a diminution of productivity and a rise of possible work-related accidents with an aging workforce [2]. Supernumerary Robotics Limbs have the potential to revolutionize industrial and manual work by helping holding heavy objects, supporting user's own arm, reaching unreachable objects or acting as a co-worker. Unlike exoskeletons that are attached to human limbs and are design to enhance existing human capabilities, a SRL is kinematically independent of user skeletal structure, which allows the device to actively perform tasks similar to or beyond human capabilities.

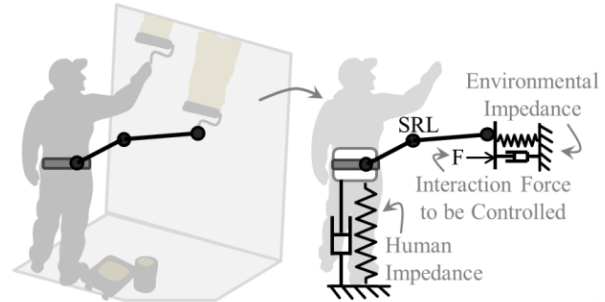

Fig. 1. SRL performing in a context of painting task. The device has to interact with human and environmental impedances.

SRLs are a recent subcategory of robotic arms and only a few research prototypes have been developed at this time. For instance, Parietti and Asada [3], [4] have developed supernumerary robotic arms for aircraft assembly, a team of the University of Tokyo [5] designed another concept of supernumerary arms to facilitate cooperation between two remote persons and MIT Media Lab developed a supernumerary robotic hand for multi-purposes domestic tasks [6].

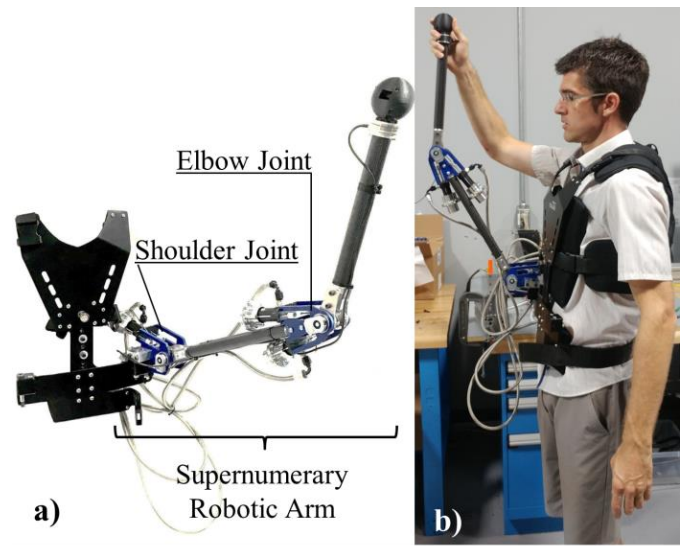

Fig. 2. Proposed 2-DOF planar wearable supernumerary robotic arm. b) Picture of the proposed wearable supernumerary arm. c) Proposed wearable supernumerary arm mounted on a user.

This work was supported by the Fonds de recherche du Québec – Nature et technologies (FRQNT), the "Ministère de l'Économie, de la Science et de l'Innovation (MESI)", Exonetik (http://www.exonetik.com) and the Natural Sciences and Engineering Research Council of Canada (NSERC).
[1] Authors are with the Faculty of Engineering, Interdisciplinary Institute for Technological Innovation (3IT), 3000 boul. de l'Université, Sherbrooke, Canada, J1K 0A5, {Catherine.Veronneau2, Jeff.Denis, Louis-Philippe.Lebel2, Alexandre.Girard2}@usherbrooke.ca.

[2] M. Denninger and J-S Plante are with Exonetik Inc., 400 Rue Marquette, Sherbrooke, Canada, J1H 1M4, {marc.denninger, js.plante}@exonetik.com.

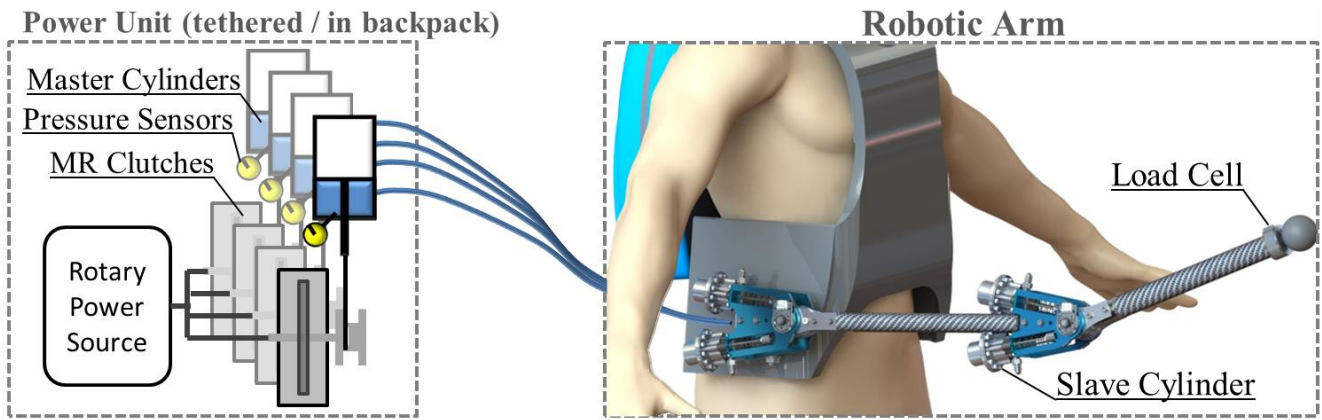

Fig. 3. Proposed concept interacting with an external impedance. The entire device is composed of the power unit and the robotic arm.

With the aim of helping the user performing his or her daily industrial tasks, like painting on a wall, grinding or machining, the SRL have to control interaction forces between the end-effector and the environment (Fig. 1). The ability of controlling interaction forces with the environment is even more challenging since the robotic device is attached on the user instead of being fixed. User's motions induce unpredictable disturbances that must be compensated by the system's controller capable of operating at high-bandwidth while being stable. To help the controller achieve those requirements, the mechanical system (structure, actuators and transmission) needs to be as mechanically transparent as possible, meaning, having low frictional and backdriving forces. Furthermore, for human-robot interaction, frictional and backdriving resistive forces must naturally be as low as possible to not constraint human intended motions [7]. Overall force bandwidth should be at least more than human force bandwidth (10 Hz) to update force response faster than the user [8].

Existing actuation systems used in wearable robotic are not suitable for SRLs applications in terms of their mechanical bandwidth, backdrivability and force-density. Traditional geared motors are widely used in classical or wearable robotic [9] but are not backdrivable nor force-controllable because of their high level of reflected inertia, friction and backlash. Direct-drive [10] or low-reduction ratio actuators are backdrivable but have a poor force-density (ref) making them unsuitable for wearable robotic. Even actuators can be used tethered to minimize total weight worn by the user, transmissions from the power unit to the slave device need to minimize additional friction or reflected inertia while remaining stiff enough to maximize bandwidth and be backdrivable. Cable-driven transmissions [11] have a low reflected inertia and a high stiffness but suffer from friction due to the cable rooting through pulleys. Pneumatic transmissions [12] avoid routing friction but suffer from a low bandwidth and a poor force-density. The challenge of developing SRLs then lies on designing a system that is intrinsically backdrivable, high-bandwidth and force-controllable while remaining enough lightweight to be worn by a user all day long.

An alternative actuation approach consisting of using magnetorheological (MR) clutches combined to a low-friction hydrostatic transmission has shown good potential for high-performance and lightweight haptic systems with a high level of backdrivability (<2-11% of resistive force on the maximum force), a high force-bandwidth (>40 Hz) and a high force-density with a remotely mounted power unit that can be detached from the robotic arm [13]. The approach was demonstrated on a haptic joystick demonstrator.

This paper presents a feasibility study evaluating the potential of the MR-Hydrostatic actuation system for a SRL application in terms of the total mass worn by the user, the backdrivability, the bandwidth and the ability to control the interaction forces with the environment. To do so, a MR-Hydrostatic 2-DOF planar supernumerary robotic arm is developed (Fig. 2). The prototype is designed to minimize total mass worn by a user, friction and reflected inertia, in order to maximize bandwidth and the ability to control interaction forces. A one-DOF test bench of the actuation system interacting with an external load is developed to measure bandwidth and compare three force-control strategies which are compared with predictions of an analytical model.

## II. MECHANICAL DESIGN OVERVIEW

The proposed supernumerary robotic arm is an additional wearable robotic arm attached on user's hip. This 2-DOF planar third arm is powered by a MR-Hydrostatic actuation [13]. Shoulder joint is designed to deliver 39 Nm with 115° of range of motion and elbow joint, 25 Nm with 180° of range of motion.

**Principle**: The whole robotic device is composed of the robotic arm and the power unit (Fig.3). The power unit contains a rotary power source (here, a geared electric motor), four MR clutches and four master cylinders. The output torque of the MR clutches is controlled by varying the current feeding an electromagnet controlling the magnetic field strength in the fluid. Similarly to an automotive brake system, each clutch winds a cable around a pulley that pulls on a master cylinder to increase hydrostatic pressure. Resulting hydraulic flow is transmitted, through a hydraulic hose, to a slave cylinder (Fig. 4) pulling on the joint cables to rotate the shoulder or elbow joint (Fig. 5). Slave cylinders stroke is 44 mm and the effective piston area is 671 $mm^2$. Linear ball-bearings are used to guide the rolling diaphragms to avoid membrane jamming. Hydrostatic transmission is filled with tap water for its high bulk modulus and ease of use.

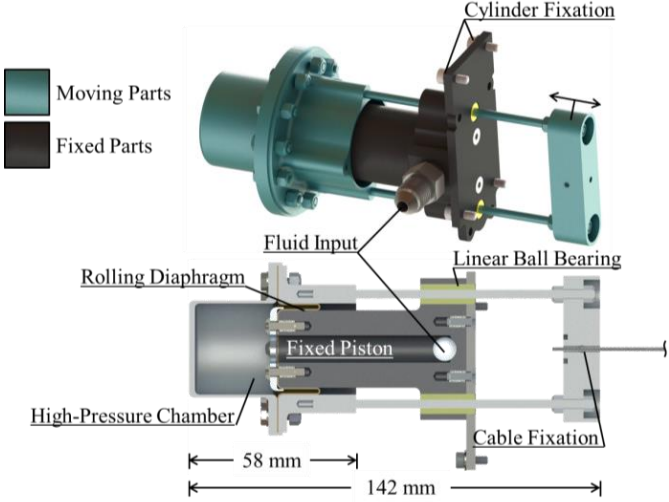

Fig. 4. Rendering of a slave cylinder used in the SRL.

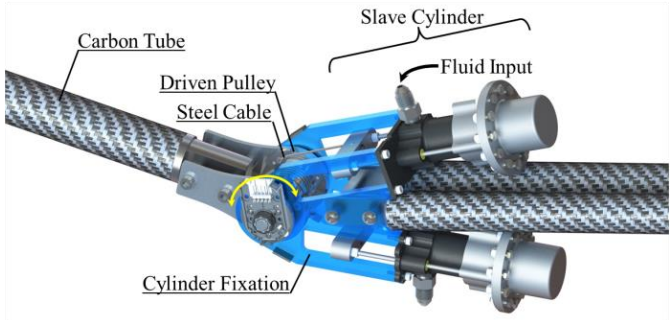

Fig. 5. Rendering of the elbow joint. Two slave cylinders mounted in an antagonist configuration and pulling on a short steel cable fixed on a driven pulley produce the rotation of the joint.

**Mass**: A significant advantage of the hydrostatic approach is that the power unit can be either mounted on the user's back for optimal mobility or be tethered to minimize the total weight worn by the user. Here, a tethered configuration is used and the robotic arm is located on the user's hips [14] to minimize the arm's total inertia perceived by the user. The robotic arm structure is also built with lightweight carbon tubes. When filled with water, the arm's total mass is 2.7 kg.

**Backdrivability**: The MR-Hydrostatic actuation system ensures a high level of backdrivability because of its low level of friction and reflected inertia [13]. First, friction in master and slave cylinders is significantly reduced by using custom-made rolling diaphragms cylinders (Fig. 4) instead of conventional cylinders [15]. Rolling diaphragm membranes roll from the bore to the piston thereby eliminating sliding motion and stick-slip friction. Second, a hydrostatic system has been chosen over a cable-driven transmission to avoid rooting and cable friction issues. Hydrostatic transmission hose internal diameter and length are also chosen to minimize fluid friction and reflected fluid inertia. For a constant cylinder effective area, the hydraulic inertia is inversely proportional to the square of the internal hose diameter and directly proportional to the hose length. Third, reflected inertia is also decreased by using MR clutches since the inertia from the electric motor is not reflected to the MR clutch output, which makes MR clutches intrinsically backdrivable actuators [16].

## III. Dynamic Performances

A one-axis, multiple degrees-of-freedom, lumped-parameter model of the MR-Hydrostatic actuation system is developed in order to: 1) predict the force-bandwidth as a function of the design parameters and 2) explore and develop closed-loop control schemes.

### A. Analytical Model

Fig. 6 presents the proposed dynamic model for the actuation system, from the current in the MR clutch to the interaction force with a load. As it will be discussed in sec. IV, this lumped-parameter with three internal DOFs is representative of the system dynamic behavior up to a frequency range of 100 Hz. The variable $x_1$ represents the reflected translation of the MR clutch output rotor, the variable $x_2$ represents the displacement of the hydraulic fluid and the variable $x_3$ represents the displacement of the output and the load. In terms of inertias, mass $m_1$ represents the reflected mass of the MR clutch output rotor and the master cylinder moving mass, $m_2$ represents the reflected mass of the hydraulic fluid, and $m_3$, the combined mass of the output assembly, slave cylinder moving mass and external load. Parameters $k_1$ and $k_2$ represent the combined compliance of the cable and the hydraulic volume in the piston (due to the flexibility of the membrane and air present in the hydraulic circuit) for respectively the power-unit side and the slave side. Parameters $b_1$, $b_2$ and $b_3$ represent viscous friction in the MR clutch, the hydraulic circuit, and the external load. Table 1 gives numerical value for all parameters, expressed in terms of reflected linear displacement at the output. Hydrostatic variables ($m_2$, $k_1$, $k_2$ and $b_2$) have been evaluated such as proposed in [13]. Values of $m_3$, $k_3$ and $b_3$ depends on the external load connected to the system.

TABLE 1.

MR-Hydrostatic Actuation Line Parameters

| | | |
|---|---|---|
| $m_1$ | MR clutch reflected mass + master cylinder piston mass | 1 kg |
| $k_1$ | Power unit transmission stiffness | $2.76 \times 10^5$ N/m |
| $b_1$ | MR clutch reflected viscous friction | 210 Ns/m |
| $m_2$ | Hydraulic fluid reflected mass | 9.65 kg |
| $k_2$ | Hydraulic stiffness | $2.76 \times 10^5$ N/m |
| $b_2$ | Hydraulic viscous damping | 98 Ns/m |
| $\tau$ | Time constant of the magnetic circuit in the MR clutch | 10 ms [17] |

The input of the model is the current input in the MR clutch which produces a torque $T_{MR}$ with a first order response of a low-pass filter representing the eddy-current limited magnetic build-up in the MR-clutch [17] with $\tau$ as time-constant. Two transfer functions are considered for the design of force feedback controllers: (1) the transfer function $H_F(s)$ between the input clutch current ($I(s)$) and the end-effector force ($F(s)$), and (2) the transfer function $H_P(s)$ between the input

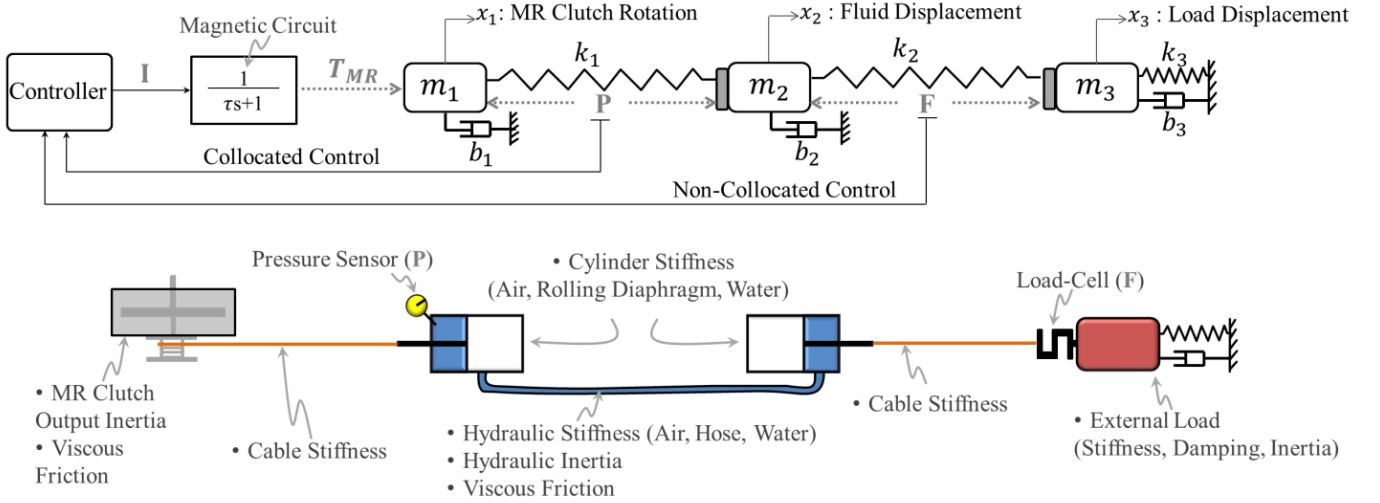

Fig. 6. One-axis lumped-parameter model of the MR-Hydrostatic actuation, from the MR clutch input current to the end effector force.

clutch current and the pressure in the hydraulic circuit (P(s)) near the master cylinder:

$$H_F(s) = \frac{F(s)}{I(s)} = \frac{1}{\tau s + 1} \cdot \frac{A(s)}{B(s)D(s) + 1} \quad (1)$$

$$H_P(s) = \frac{P(s)}{I(s)} = \frac{1}{\tau s + 1} \cdot \frac{A(s)C(s)}{B(s)D(s) + 1} \quad (2)$$

with,

$$A(s) = \frac{k_1}{m_1 s^2 + b_1 s + k_1} \quad (3)$$

$$B(s) = \frac{(m_1 s^2 + b_1 s)k_1}{m_1 s^2 + b_1 s + k_1} \quad (4)$$

$$C(s) = \frac{Z_3 k_2}{(m_2 s^2 + b_2 s)(Z_3 + k_2) + Z_3 k_2} \quad (5)$$

$$D(s) = \frac{Z_3 + k_2}{(m_2 s^2 + b_2 s)(Z_3 + k_2) + Z_3 k_2} \quad (6)$$

$$Z_3(s) = (m_3 s^2 + b_3 s + k_3) \quad (7)$$

Note that those transfer functions depend on $Z_3$ representing the impedance of the external load.

**Force Bandwidth**: The open-loop force-bandwidth is directly related to frequency response of $H_F(s)$ that describes the behavior of the interaction force responding to an input current in the clutch. $H_F(s)$ is analyzed for two scenarios (Fig. 10 and Fig. 11 with analytical results), when the output is 1) blocked ($Z_3(s) \to \infty$) and 2) connected to a compliant load ($k_3$ =12 000 N/m, $b_3$ =20 Ns/m and $m_3$ =1.87 kg) that represents the presence of human and/or environmental impedance (see Fig. 1). When the output is blocked, the total system force bandwidth is 25.4 Hz, and decreases to 6.5 Hz with the compliant load. Thus, due to low intrinsic impedance of the actuator and transmission, open-loop force control can be sufficient for many situations. However, if the task is dynamic, a closed-loop approach can be necessary to achieve a better control of the interaction force, especially if the impedance of the environment/human is low.

**Force Feedback**: Two closed-loop approaches are explored for the proposed robotic arm. The first traditional approach is to have a load-cell at the output (arm end effector), that measures directly the force to be controlled. This approach can work very well for quasi-static task but have fundamental issues due to the non-collocated force feedback [18]. By analyzing $H_F(s)$, it is possible to see that large feedback gain will lead to instability before a very high closed-loop bandwidth can be achieved (see root locus Fig. 7 and the bode plots of Fig. 10 and Fig. 11). An alternative approach that is possible with the proposed architecture is to measure the pressure in the hydraulic line to close the loop. This approach is advantageous from a hardware point of view because a load-cell is much more expensive and heavier than pressure sensors. Furthermore, the pressure feedback is closer to be collocated with the force source, and thus inherently more stable. By analyzing $H_P(s)$, it is possible to see that this closed-loop is more stable and has the potential for a higher closed-loop bandwidth with all poles in the left plane and a phase that stays within safe bounds (see root locus Fig. 7 and the bode plots of Fig. 10 and Fig. 11).

Fig. 8 illustrates three force-controls approaches that are possible with a MR-Hydrostatic actuation. The open-loop force-control (Fig. 8a) transforms the reference force ($F_{ref}$) to an input current command (I) to the MR clutch with a static gain ($G_1$). In the closed-loop force-control with the end effector force feedback (Fig. 8b), the controller takes the error between the reference force and the force measured by the end effector load-cell ($F_{meas}$). This non-collocated control requires adding a load-cell force-control to the arm's end effector. The closed-loop force-control with the master cylinder pressure feedback (Fig. 8c) compares a reference pressure ($P_{ref}$) to the measured pressure ($P_{meas}$). The reference pressure is obtained by multiplying the reference force by a static gain ($G_2$). This collocated force-control offers a cheaper option by avoiding the utilization of expensive force sensors on the robotic arm and also decreases total mass on the arm.

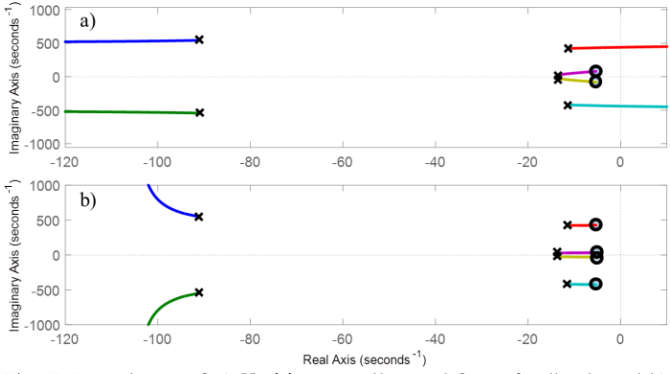

Fig. 7. Root locus of a) $H_F(s)$ non-collocated force-feedback and b) $H_P(s)$ pressure feedback. Feedback gains are fundamentally much more limited for $H_F(s)$ than $H_P(s)$.

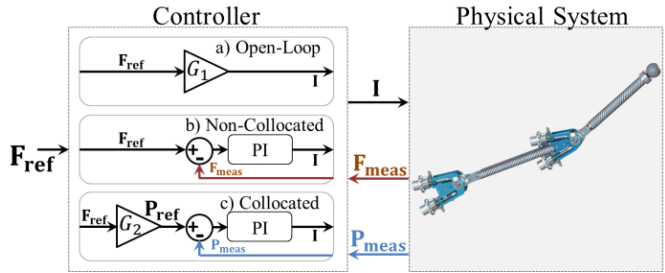

Fig. 8. Block diagrams the three force-control strategies that are possible to implement with a MR-Hydrostatic actuation; 1) the open-loop force-control, 2) the closed-loop force-control with force feedback and 3) the closed-loop force-control with master cylinder pressure feedback.

## IV. Experimental Results

This section evaluates the dynamic performances of the actuation system and validates the proposed analytical model. The 2-DOFs prototype shown at Fig. **2** is used for a realistic application demonstration and a one-DOF test-bench of the MR-Hydrostatic actuation is used for performance characterization tests: an experimental evaluation of the open-loop force-bandwidth and output force tracking tests with the three control approaches.

### A. One-DOF MR-Hydrostatic Test Bench

Fig. 9 presents the power unit coupled through the hydrostatic transmission to the slave part of the test bench. The power unit is composed of one electric geared motor (HC775LP-301 Johnson Electric with a 25:1 gearbox) and a 4 Nm MR clutch pulling on a master cylinder through a 12 mm diameter pulley and a cable. A current drive (30A8 Analog, Advanced Motion Control) regulates input current to the MR clutch. The custom-made master cylinder effective area is 826 mm$^2$ and is made with a rolling diaphragm (4-137-119-BBJ, Marsh Bellofram) rated for a maximum pressure of 3 100 kPa. The pressure in the master cylinder is monitored by a pressure sensor (PX3AN1BH667PSAAX, Honeywell). A nylon-braided hydraulic hose with a 9.5 mm internal diameter and filled with tap water is used as the hydraulic transmission. Fluid pressure is then transmitted to a custom-made slave cylinder of effective area of 671 mm$^2$ and made with a rolling diaphragm (4-125-106-BBJ, Marsh Bellofram). A 60 mm long steel cable (#2081, Carl Stahl) of 1.98 mm of diameter is used to pull an external impedance composed of a mass (1.87 kg), a spring (12 000 N/m) and a damper (20 Ns/m). A load-cell (SBO-200, Transducer Techniques) measures the interaction force between the actuation line and the impedance. The complete test bench is controlled by a real-time *Speedgoat* controller at a sampling rate of 1.5 kHz.

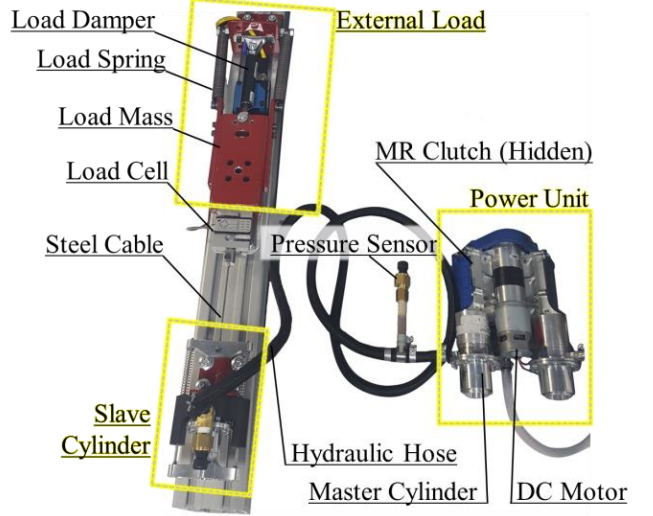

Fig. 9. One-DOF experimental MR-Hydrostatic test bench designed and built to evaluate the force bandwidth and to test three force-control strategies.

### B. Open-Loop Force-Bandwidth

The following section presents the experimental open-loop force-bandwidth of the MR-Hydrostatic actuation line tested with the one-DOF test bench. A current input command ranging from 1 to 3.5 A and from 0 to 100 Hz (logarithmic chirp) is sent to the MR clutch. The experimental equivalent open-loop transfer functions $H_F(s)$ and $H_P(s)$ are obtained and compared to the analytical model for two output conditions: blocked (Fig. 10) or connected to a compliant load (Fig. 11). The proposed model is found to fit closely experimental results in both output conditions. Only the damping parameters ($b_1$, $b_2$, and $b_3$) were adjusted to fit the experimental frequency responses. The other model parameters were evaluated analytically (mass $m_1$, $m_2$, and stiffness $k_2$) or measured experimentally (mass $m_3$ and stiffness $k_1$, and $k_3$). The proposed analytical model is thus a good tool for design and performance prediction of this type of actuation system, expect for the damping ratios that are harder to predict analytically.

### C. Closed-Loop Force-Control

To control the interaction forces acting on the arm's end-effector, three force-controls are compared (Fig. 8): (1) open-loop control, (2) closed-loop control with force feedback and (3) closed-loop control with master cylinder pressure feedback. Those tests are conducted with the one-DOF test bench interacting with the compliant external load. To compare these approaches, a mixed force reference signal including a step and a chirp is send to the force controller. The target force input command begins with a step from 50 N to 250 N followed by a logarithmic chirp with a 100 N amplitude sweeping from 0 to 6 Hz. Both closed-loop approaches use an experimentally tuned PI controller.

Fig. 12 shows that both closed-loop approaches lead to better tracking of the reference signal, compared the open-loop approach where the force lag behind the reference. Rise times are also improved with the closed-loop approaches: 56 ms and 70 ms, compared to 81 ms in open-loop. Regarding the two closed-loop approaches, the two lead here to similar tracking performances in this experiment. However, with the load-cell force-feedback the system is closer to instability. The barely damped pair of poles that can cross-over the right-hand plane (Fig. 7) with high gain is easily excited and can add undesirable high-frequency oscillations during transients as seen at Fig. 12b.

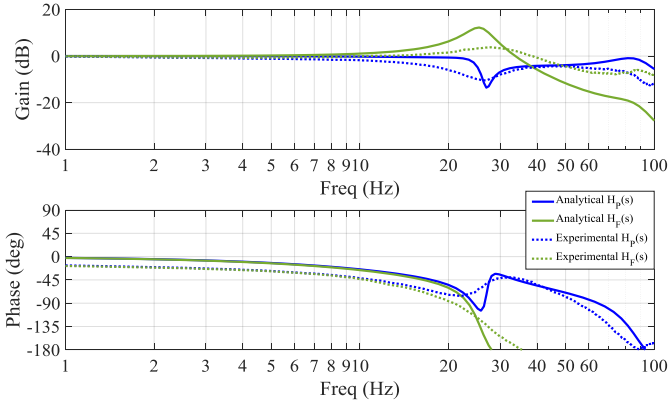

Fig. 10. Analytical and experimental Bode plots of $H_P(s)$ and $H_F(s)$ evaluated for the blocked output condition.

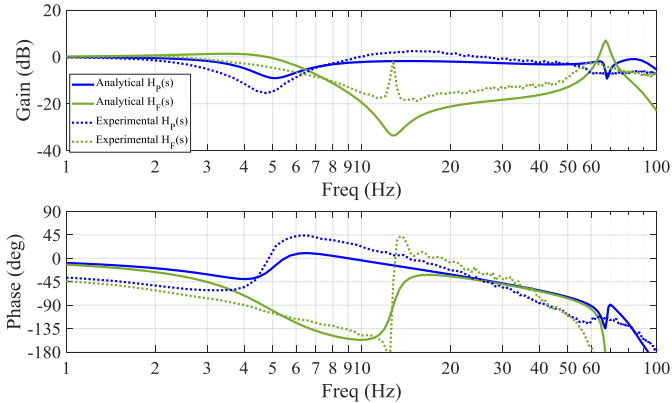

Fig. 11. Analytical and experimental Bode plots of $H_P(s)$ and $H_F(s)$ evaluated when output is connected to a compliant load.

### D. Application Demonstration

Fig. 13 presents an industrial demonstration of one of the possible functionalities of the arm. In this demonstration, a user puts a wood panel on a wall with her two own arms and the robotic arm then stabilize the panel during a drilling task, by controlling the interaction force normal to the wall in order to keep the panel in place. The force is recorded with the end-effector load cell and a controller tracks a constant reference force of 23 N (approach b: load-cell force feedback). Deviations of the force during the drilling task are roughly ±2 N (shown by highlighted region).

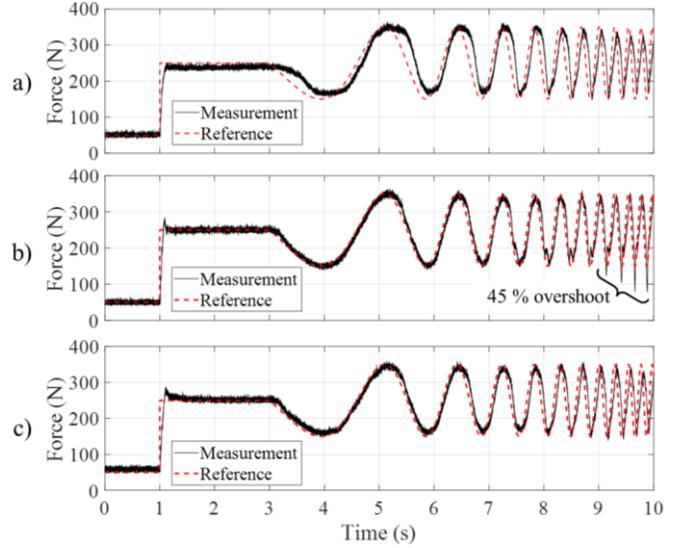

Fig. 12. a) open-loop force-control, b) closed-loop force-control with force feedback and c) closed-loop force-control with master cylinder pressure feedback.

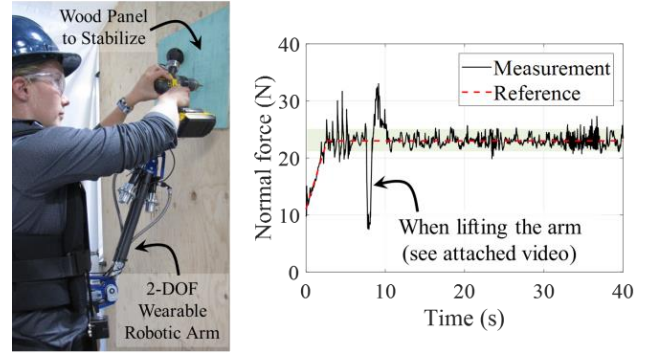

Fig. 13. Industrial demonstration of the proposed supernumerary robotic arm with a force-tracking.

## V. CONCLUSION

This paper studied the feasibility of using MR clutches coupled to hydrostatic lines and rolling diaphragm cylinders as an actuation strategy that can provide high interaction dynamics while remaining lightweight for SRL applications.

Analytical and experimental results from a one-DOF test bench show that the approach has sufficient open-loop bandwidth for wearable human interaction with 25 Hz when blocked and 6 Hz when coupled to an impedance. Three force control approaches are demonstrated: open-loop, closed-loop on force, and closed-loop on pressure. The pressure approach is particularly interesting because it is collocated and has better stability at high frequency.

Finally, a 2-DOFs wearable robotic arm demonstrator is built showing a mass of 2.7 kg that is lightweight enough to be worn for long periods of time with no strain. The arm is demonstrated in a panel installation application and shows that a relatively constant force can be applied on the work object while the human exhibits natural motion.

In all, although future work is needed to mature the proposed technology, the results presented in this paper suggest the MR-hydrostatic approach to be a promising actuation technology for SRL systems.